\documentclass[runningheads]{llncs}
\usepackage{graphicx}
\usepackage{amsmath}
\usepackage{amssymb}
\usepackage{algorithm}
\usepackage{algorithmic}
\usepackage{subfig}
\usepackage{placeins}
\usepackage{xcolor}
\usepackage{url}
\usepackage{breakurl}
\begin{document}
\title{Automatic Setting of DNN Hyper-Parameters by Mixing Bayesian Optimization and Tuning Rules}
\titlerunning{DNN-Tuner}
%
\author{Michele Fraccaroli\inst{1} \orcidID{0000-0001-7656-7204} \and
Evelina Lamma\inst{1} \orcidID{0000-0003-2747-4292} \and
Fabrizio Riguzzi\inst{2}\orcidID{0000-0003-1654-9703} }
\authorrunning{M. Fraccaroli et al.}
%
\institute{DE - Department of Engineering \and
DMI - Department of Mathematics and Computer Science \\
University of Ferrara\\ 
Via Saragat 1, 44122 Ferrara, Italy \\
\email{\{michele.fraccaroli, evelina.lamma, fabrizio.riguzzi\}@unife.it}}
\maketitle              
\begin{abstract}
Deep learning techniques play an increasingly important role in industrial and research environments due to their outstanding results. However, the large number of hyper-parameters to be set may lead to errors if they are set manually. The state-of-the-art hyper-parameters tuning methods are grid search, random search, and Bayesian Optimization. The first two methods are expensive because they try, respectively, all possible combinations and random combinations of hyper-parameters. Bayesian Optimization, instead, builds a surrogate model of the objective function, quantifies the uncertainty in the surrogate using Gaussian Process Regression and uses an acquisition function to decide where to sample the new set of hyper-parameters. This work faces the field of  Hyper-Parameters Optimization (HPO). The aim is to improve Bayesian Optimization applied to Deep Neural Networks. For this goal, we build a new algorithm for evaluating and analyzing the results of the network on the training and validation sets and use a set of tuning rules to add new hyper-parameters and/or to reduce the hyper-parameter search space to select a better combination.

\end{abstract}
\section{Introduction}
\label{intro}
Deep Neural Networks (DNNs) provide outstanding results in many fields but, unfortunately, they are also very sensitive to the tuning of their hyper-parameters. Automatic hyper-parameters optimization algorithms have recently shown good performance and, in some cases, results comparable with human experts \cite{bergstra2011algorithms}.
Tuning a big DNN is computationally expensive and some experts still perform manual tuning of the hyper-parameters. To do this, one can refer to some tricks \cite{montavon2012neural} to determine the best set of hyper-parameter values to use for obtaining good performance from the network.
This work aims to combine the automatic approach with these tricks used in the manual approach but in a fully-automated integration for drive the training of DNNs, automatizing the choice of HPs and analyzing the performance of each training experiment to obtain a network with better performance. Heuristics and Bayesian Optimization for parameter tuning have already been used in other application contexts other than Deep Learning, e.g., using heuristics to shrink the size of a parameter space for Self-Adapting Numerical Software (SANS) systems \cite{dongarra2003self}, using rollout heuristics that work well with Bayesian optimization when a
finite budget of total evaluations is prescribed  \cite{lam2016bayesian} or using domain-specific knowledge to reduce the dimensionality of Bayesian optimization to tune a robot's locomotion controllers \cite{rai2018bayesian}.
For the automatic approach,  we use a Bayesian Optimization \cite{dewancker2015bayesian} because it limits the evaluations of the objective function (training and validation of the neural network in our case) by spending more time in choosing the next set of hyper-parameter values to try.
We aim at improving Bayesian Optimization \cite{dewancker2015bayesian} by integrating it with a set of tuning rules. These rules have the purpose of reducing the hyper-parameter space or adding new hyper-parameters to set. In this way, we constrain the Bayesian approach to select the hyper-parameters in a restricted space and add new parameters without any human intervention. In this way, we can avoid typical problems of the neural networks like overfitting and underfitting, and automatically drive the learning process to good results.

After discussing Bayesian Optimization (Section \ref{rw}), we introduce the approach used to analyze the behaviour of the networks, the execution pipeline, the performed analysis and the tuning rules used to fix the detected problems (Section \ref{nba}, Section \ref{EP}, Section \ref{PA}, and Section \ref{tr}). Experimental results with different networks and different datasets are described in Section \ref{ex}. All experiments are performed on benchmark datasets, MNIST and CIFAR-10 in particular.

\section{Bayesian Optimization}
\label{rw}
In this section, we shortly review the state-of-the-art of hyper-parameter optimization for DNNs, i.e., grid search, random search  and, with special attention, Bayesian Optimization.
Grid Search applies exhaustive research (also called \textit{brute-force search}) through a manually specified hyper-parameter space. This search is guided by some specified performance metrics.
Grid Search tests the learning algorithm with every combination of the hyper-parameters and, thanks to its \textit{brute-force search}, guarantees to find the optimal configuration. But, of course, this method suffers from the \textit{curse of dimensionality}.
The problem of Grid Search applied to DNNs is that we have a huge amount of hyper-parameters to set and the evaluating phase of the algorithm is very expensive. This definitively raises a time problem.

\medskip
Random Search uses the same hyper-parameter space of Grid Search, but replaces the \textit{brute-force search} with \textit{random search}. It is, therefore, possible that this approach will not find results as accurate as Grid Search, but it requires less computational time while finding a reasonably good model in most cases \cite{bergstra2012random}.

\medskip
Bayesian Optimization (BO) is a probabilistic model-based approach for optimizing objective functions ($f$) which are very expensive or slow to evaluate \cite{dewancker2015bayesian}. The key idea of this approach is to limit the evaluation phase of the objective function by spending more time to choosing the next set of hyper-parameters' values to try.  
In particular, this method is focused on solving the problem of finding the maximum of an expensive function
 $f: X \to \mathbb{R}$, 
\begin{equation}
arg max_ {x \in X}  f(x)
\end{equation}
where $X$ is the hyper-parameter space while can be seen as a hyper-cube where each dimension is a hyper-parameter.

BO builds a surrogate model of the objective function and quantifies the uncertainty in this surrogate using a regression model (e.g., \textit{Gaussian Process Regression}, see next section). Then it uses an acquisition function to decide where to sample the next set of hyper-parameter values \cite{frazier2018tutorial}.

\begin{figure}
	\centering
	\hspace*{-1.75cm}
	\subfloat[]{\includegraphics[scale=0.17]{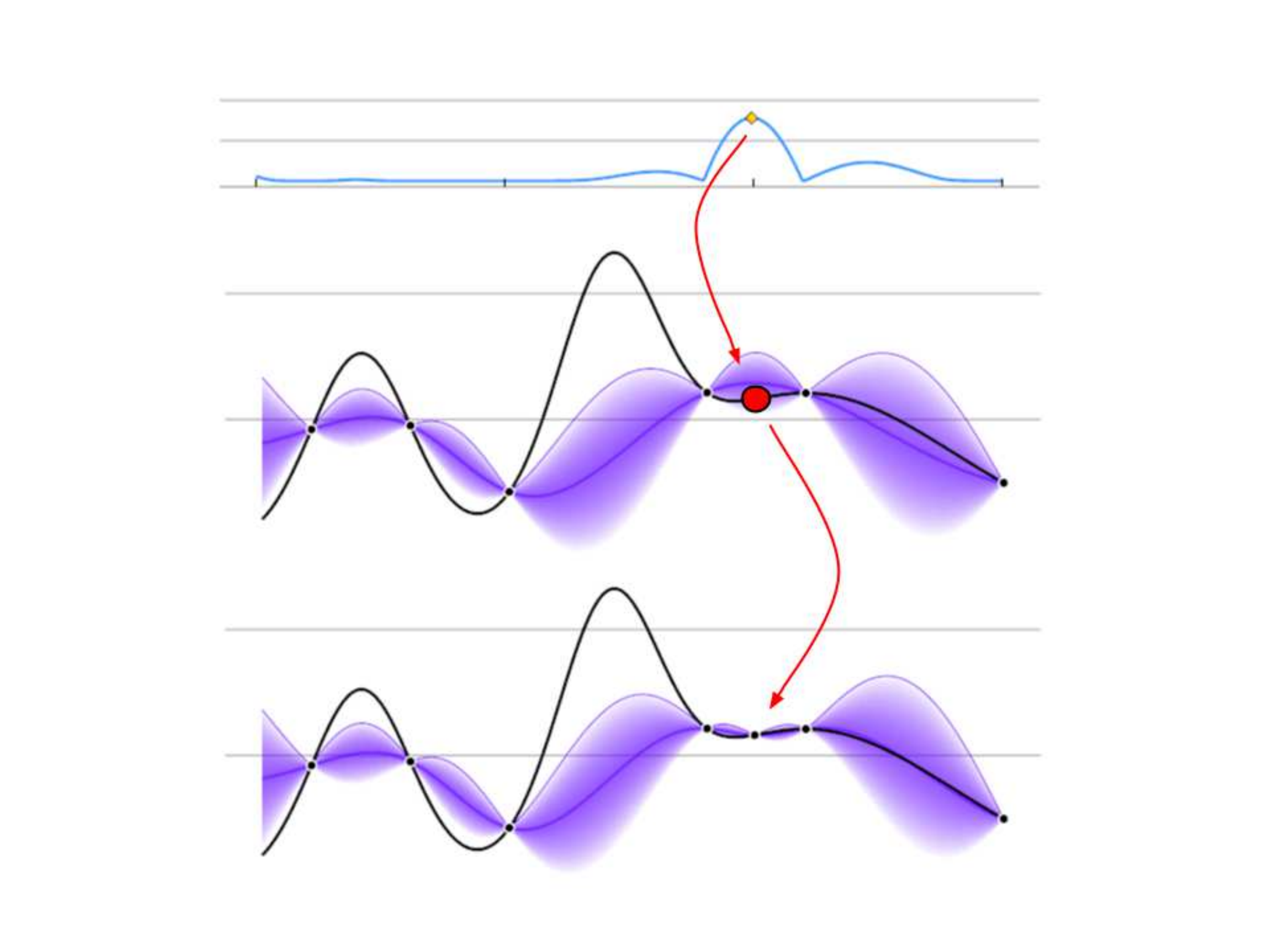}}
	\subfloat[]{\includegraphics[scale=0.4]{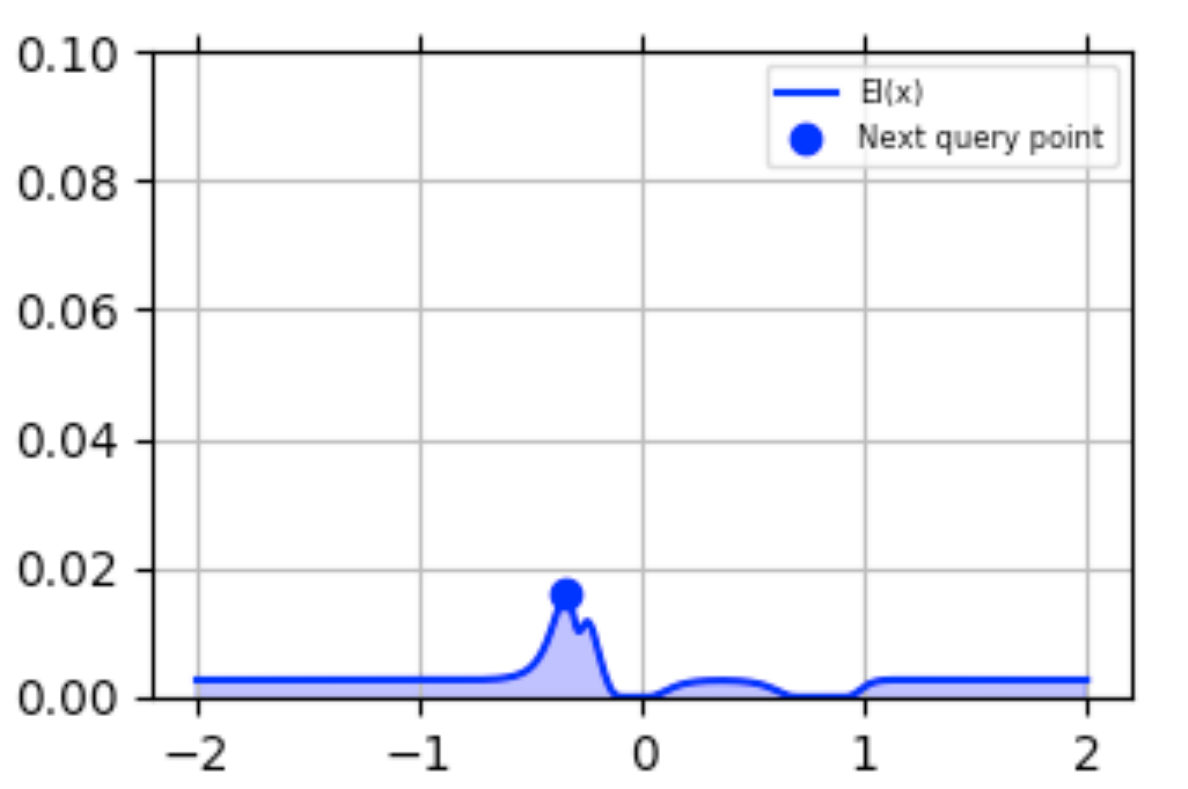}}
	\caption{(a) BO behaviour. The figure shows the objective function (black line), mean (light purple line), covariance (purple halo) and the samplings (black dots). The top of the figure shows the Expected Improvement function (blue line) and his maximum (yellow dot). The next sampled point will be exactly the maximum of the activation function (red dot). The bottom of the figure shows  the new statistical model \protect\cite{wiki:gpbayesian}. (b) Example of Expected Improvement with the sampling of the next point (blue dot) generated with Scikit-Optimize library.} 
	\label{fig: GPr}
\end{figure}

The two main components of Bayesian Optimization are: the \textit{statistical model} (regression model) and the \textit{acquisition function} for deciding the next sampling. 
The statistical model provides a posterior probability distribution that describes potential values of the objective function at candidate points. As the number of observations grows, the posterior distribution improves and the algorithm becomes more certain of which regions in hyper-parameter space are worth to be explored and which are not. The acquisition function is used to find the point where the objective function is supposed to be maximal \cite{cosmetic}. This function defines a balance between \textit{Exploration} (new areas in the parameter space) and \textit{Exploitation} (picking values in areas that are already known to have favorable values); the aim of the \textit{Exploration} phase is to select samples that shrink the search space as much as possible, and the aim of the \textit{Exploitation} phase is to focus on the reduced search space and to select samples close to the optimum \cite{DBLP:journals/corr/abs-1204-0047}. Figure \ref{fig: GPr} (a) shows the behaviour of this algorithm.

\medskip
A Gaussian Process (GP) is a stochastic process and a powerful prior distribution on functions. Any finite set of $N$ points ${\{ x_n \in X \}^{N}_{n=1}}$ induces a multivariate Gaussian distribution on $\mathbb{R}^{N}$. GP is completely determined by the mean function $m: X \to \mathbb{R}$ and the covariance function $K: X \times X \to \mathbb{R}$ \cite{rasmussen2003gaussian}.  For a complete and more precise overview of the Gaussian Process and its application to Machine Learning, see \cite{rasmussen2003gaussian}.

\medskip
The acquisition function is used to decide where to sample the new set of hyper-parameters' values, Figure \ref{fig: GPr}(b) shows. In the literature, there are many types of acquisition function. Usually, only three are used: \textit{Probability of Improvement}, \textit{Expected Improvement} and \textit{GP Upper Confidence Bound} (GP-UCB) \cite{cosmetic} \cite{snoek2012practical}.
The acquisition function used in this work is \textit{Expected Improvement} \cite{jones1998efficient}. The idea behind this acquisition function is to define a non-negative expected improvement over the best previously observed target value (previous best value) at a given point $x$. Expected Improvement is formalized as follows:
\begin{equation}
\label{EI}
EI_{y^*}(x) = \int_{-\infty}^{y^*} (y^* - y)p(y|x) \, dy
\end{equation}
where $y^*$ is the actual best result of the objective function, $x$ is the new proposed set of hyper-parameters' values, $y$ is the actual value of the objective function using $x$ as hyper-parameters' values and $p(y|x)$ is the surrogate probability model expressing the probability of $y$ given $x$.

\section{Network Behaviour Analysis}
\label{nba}
By automatically analyzing the behaviour of the network, it is possible to identify problems that DNNs could have after the training phase (e.g., overfitting, underfitting, etc).  Bayesian Optimization only works with a single metric (validation loss or accuracy, training loss or accuracy) and is not able to find problems like overfitting or fluctuating loss.
When these problems are diagnosed, the proposed algorithm aims to shrink the hyper-parameters' value space or update the network architecture to drive the training to a better solution.

Many types of analyses can be performed on the network results, e.g., analyzing the relation between training and validation loss or the accuracy, the trend of accuracy or loss during the training phase in terms of direction and form of the trend.
For example: if a significant difference between the training loss and validation loss is found, and the validation loss is greater then the training loss, the diagnosis is that the network has a high variance and overfitting occurred.

\subsection{Execution Pipeline}
\label{EP}
\begin{figure}
	\centering
	\includegraphics[scale=0.18]{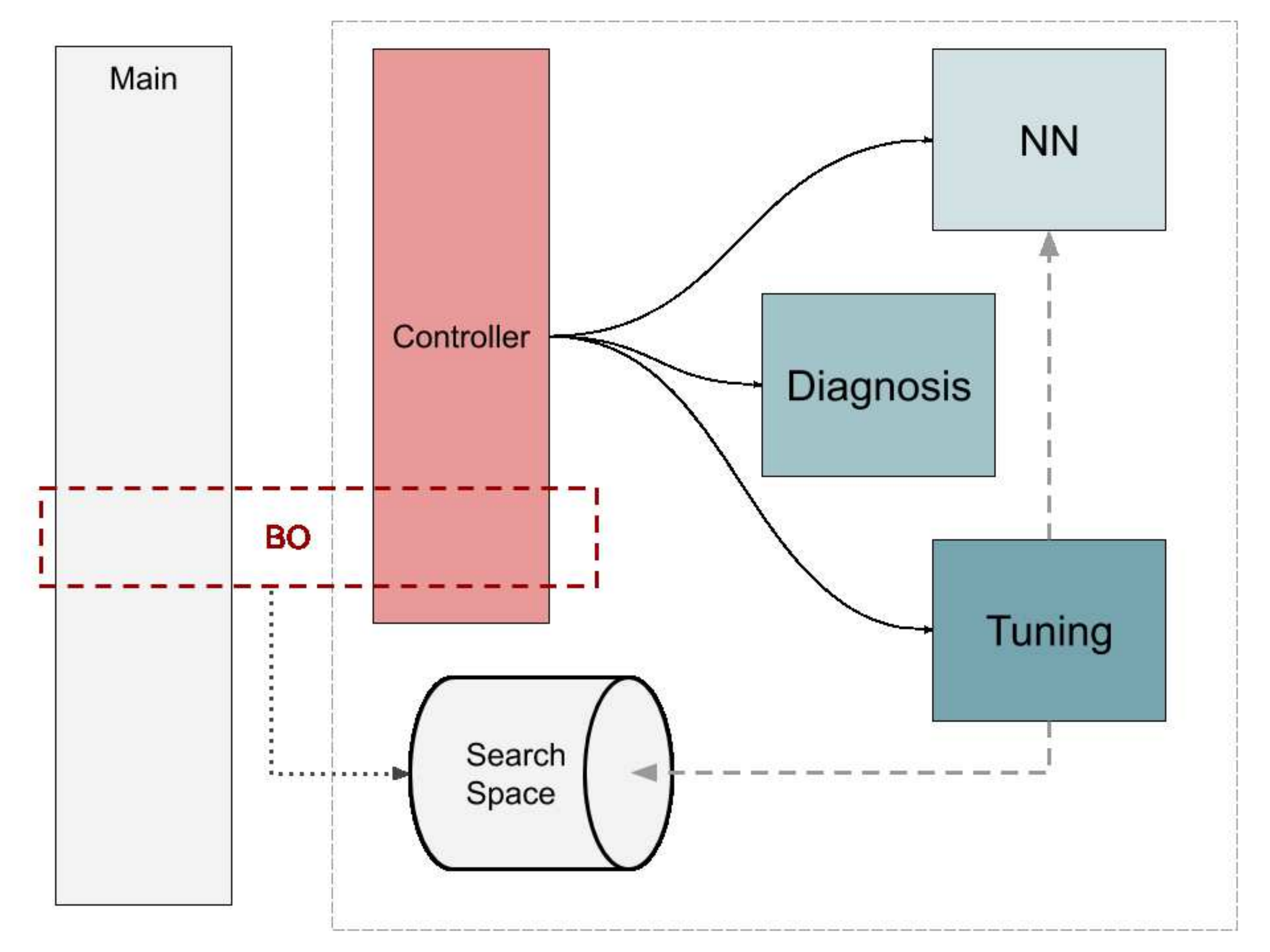}
	\caption{The structure of the software with the main modules.} \label{fig: pipeline}
\end{figure}

The software structure is composed by four main modules: \textit{training} module (NN in Figure \ref{fig: pipeline}), \textit{Diagnosis} module, \textit{Tuning Rules} module and the \textit{Search Space} driven by the \textit{controller} as can be seen in Figure \ref{fig: pipeline}. Controller rules the whole process. The main loop of the Algorithm \ref{alg: DNN-Tuner} uses the Controller for running the \textit{Diagnosis} module, the \textit{Tuning Rules} module and BO for performing the optimization. BO is used for choose the hyper-parameters' value and is forced to do so from a restricted search space.

Algorithm \ref{alg: DNN-Tuner} encapsulates the whole process (called \textit{DNN-Tuner}).

With $S$, $N$, $Params$ and $n$ we refer respectively to the hyper-parameters' value search space, initial neural network definition, a value picked from the search space and the number of cycles of the algorithm. $R$, $H$, and $M$ are the results of the evaluation of the trained network, the history of the loss and accuracy in both training and validation phases and the trained model of the network, respectively. $Ckpt$ is a checkpoint of the Bayesian Algorithm that can be used to restore the state of the Bayesian Optimization. $Issues$, $NewM$ and are a list of issues found by the \textit{Diagnosis} module, the new model, respectively. 

All the modules are implemented in Python. The DNN and BO are implemented with Keras and Scikit-Optimize respectively.

\begin{algorithm}
	\caption{DNN-Tuner}
	\begin{algorithmic}
		\REQUIRE $S$, $N$, $n$
		\STATE $Params \leftarrow Randomization(S)$
		\STATE $M \leftarrow Build Model(P,N)$
		\STATE $R,H \leftarrow Training(Params, M)$
		\STATE $Ckpt \leftarrow None$
		\WHILE{$n > 0$}
		\STATE $Issues \leftarrow Diagnosis(H, R)$
		\STATE $New M, S \leftarrow Tuning(Issues, M, S)$
		\IF{$New M \neq M$}
		\STATE $Params \leftarrow Randomization(S)$
		\STATE $R,H \leftarrow Training(Params, New M)$
		\ELSE
		\STATE $Ckpt \leftarrow BO(S,NewM, Ckpt)$
		\ENDIF
		\STATE $n \leftarrow (n-1)$
		\ENDWHILE
	\end{algorithmic}
	\label{alg: DNN-Tuner}
\end{algorithm}

\subsection{Diagnosis Module}
\label{PA}
We have performed some proof-of-concept analyses. DNN-Tuner analyses the correlation between training and validation accuracy and loss to detect overfitting. It also analyses validation loss and accuracy to detect underfitting, and the trend of loss to fix the learning rate and the batch size.

\begin{figure}
	\centering
	\hspace*{-1.05cm}
	\subfloat[]{\includegraphics[scale=0.70]{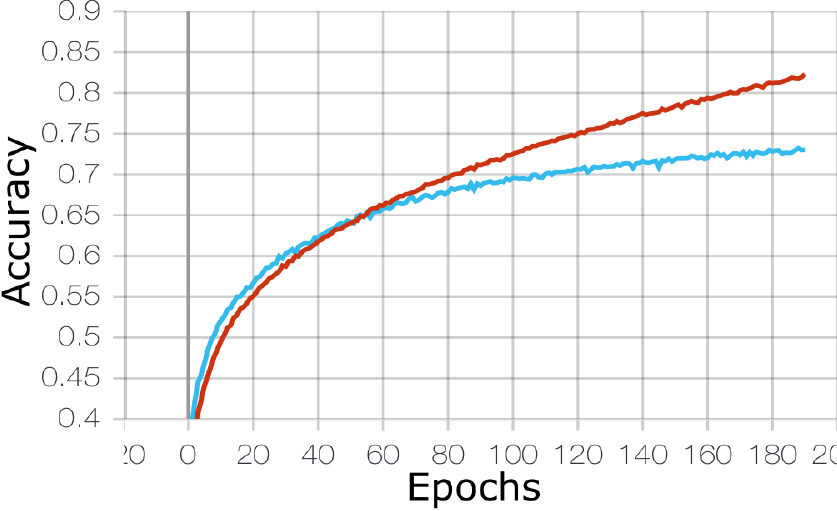}}
	\subfloat[]{\includegraphics[scale=0.70]{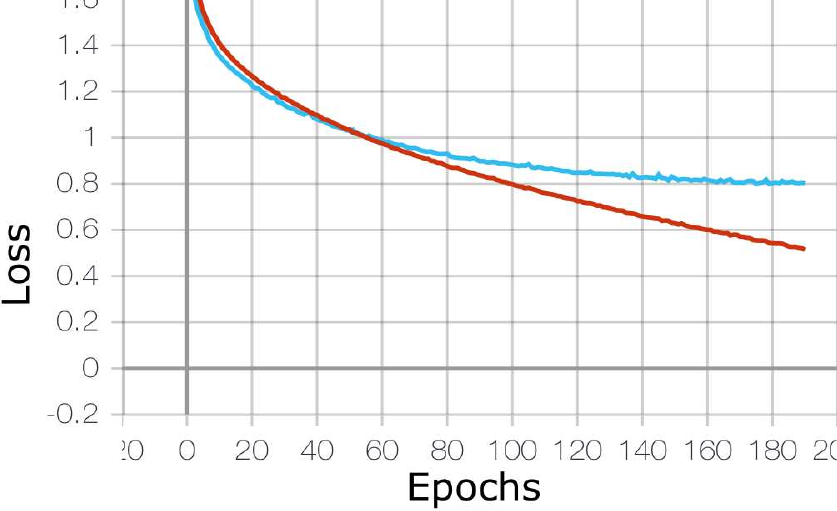}}
	\caption{An example of accuracy (a) and loss (b) in the training (red line) and validation phase (light blue line).} \label{fig: overfitting}
\end{figure}

For the first analysis, the proposed algorithm evaluates the difference of the accuracy in the training and validation phases to identify any gap as shown by Figure \ref{fig: overfitting}. An important gap between these two plots is a clear sign of overfitting since this means that the network classifies well training data but not so well validation data, that is, it is unable to generalize. In this case, the algorithm can diagnose overfitting and then take action to correct this behaviour.
\FloatBarrier

For detecting underfitting, DNN-Tuner analyses the results of loss in both phases and estimates the amount of loss. The loss value that can be used as a threshold to diagnose underfitting is variable and depends on the problem. Then, a threshold is imposed and, if this threshold is exceeded, DNN-Tuner will diagnose the underfitting.

The analysis of the trend of the loss is useful for detecting the effects of the learning rate and batch size on the network. Given the trend, the proposed algorithm can check whether the loss is too noisy, and whether the direction of the loss is descending, ascending or constant. In response to this diagnosis, the algorithm can attempt to set the batch size and learning rate correctly. 
These adjustments are performed by applying rules called \textit{Tuning Rules} (Section \ref{tr}).

\subsection{Tuning Rules}
\label{tr}
\textit{Tuning rules} are the actions applied to the network or to the hyper-parameters' value search space in response to the issues identified with the \textit{Diagnosis} module.

\medskip
Then, for each issue found by the diagnosis module, a tuning rule is applied and a new search space and a new model are derived from the previous ones.
In the context of this work, the rules applied for each issue are shown Table \ref{table: rules}.

\begin{table}[h!]
	\centering
	\begin{tabular}{| l | l |}
		\hline
		Issue & Tuning Rules\\ [1ex] 
		\hline
		Overfitting & Regularization \\ & Batch Normalization \\
		Underfitting & More neurons \\
		Fluctuating loss & Increase of batch size \\
		Increasing loss trend & Decrease of learning rate \\
		\hline
	\end{tabular}
	\caption{Rules applied on each issue.}
	\label{table: rules}
\end{table}
In order to prevent overfitting, the tuning module applies Batch Normalization \cite{DBLP:journals/corr/IoffeS15} and L2 regularization \cite{DBLP:journals/corr/Laarhoven17b}. For underfitting, it tries to increase the learning capacity of the network by removing small values from the space of the number of neurons in the convolutional and the fully connected layers. In this way, it drives the optimization to choose higher values for the number of neurons. If DNN-Tuner finds the trend of the loss to be fluctuating, it forces the optimization to choose larger batch size values.


The amount of oscillation in the loss is related to the batch size. When the batch size is 1, the oscillation will be relatively high and the loss will be noisy. When the batch size is the complete dataset, the oscillation will be minimal because each update of the gradient should improve the loss function monotonically (unless the learning rate is set too high). 
The last rule of Table \ref{table: rules} tries to fix a growing loss. When the loss grows or remains constant at too high values, this means that the learning rate is probably too high and the learning algorithm is unable to find the minimum because it takes too large steps. When the algorithm detects this behaviour, it removes large values from the learning rate search space.

\medskip
Tuning is applied following the IF-THEN paradigm as shown in Algorithm \ref{alg: tuning}, where $NewS$ is the altered search space.
\begin{algorithm}
	\caption{Tuning}
	\begin{algorithmic}
		\REQUIRE $Issues, M, S$
		\FOR{$I$ in $Issue$}
		\IF{$I = ``overfitting"$}
		\STATE $NewM, NewS \leftarrow Apply([Regularization, Batch\ Normalization], M, S)$
		\ELSIF{$I = ``underfitting"$}
		\STATE $NewM, NewS \leftarrow Apply(More\ neurons, M, S)$
		\STATE ...
		\ENDIF
		\ENDFOR
		\RETURN $NewM, NewS$
	\end{algorithmic}
	\label{alg: tuning}
\end{algorithm}

\section{Experiments}
\label{ex}
We compare \textit{DNN-Tuner} with standard Bayesian Optimization implemented with the library Scikit-Optimize. Experiments  were performed on CIFAR-10 and MNIST dataset. For each experiment, a cluster with NVIDIA GPU K80 and 8-cores Intel Xeon E5-2630 v3 was used. Due to the hardware settings, there is a time limit of 8 hours for each the experiment.

The initial state of the DNN is composed of two blocks consisting of two Convolutional layers with ReLU as activation followed by a MaxPooling layer. At the end of the second block, there is a Dropout layer followed by a chain of two Fully Connected layers separated by a Dropout layer.
The initial hyper-parameters to be set by the algorithm are the number of the neurons in the Convolutional layers, the values of the Dropout layers, the learning rate and the batch size.
The size of the search space depends on hyper-parameter. The domains of hyperparameters are as follows.
The size of the first and second convolutional layers are between 16 and 48 and between 64 and128 respectively. The domains of the rate of dropout for the first and second convolutional layers are [0.002,0.3] and [0.03,0.5] respectively.
The size of the fully connected layes is between 256 and 512. The domain of the learning rate is [$10^{-5}$,$10^{-1}$].

Unlike Bayesian Optimization, our algorithm can update the network structure and the hyper-parameters to be set. For example, when our algorithm detects overfitting, it updates the network structure by adding L2 regularization at each Convolutional layers (kernel regularization), Batch Normalization after the activations and adds L2 regularization parameters to the hyper-parameters to be set. Both DNN-Tuner and BO start with the same neural network and the same hyper-parameters to tune  and the same domains for hyperparameters..

\begin{figure}
	\centering
	\subfloat[Accuracy of DNN-Tuner.]{\includegraphics[width=170pt, height=80pt]{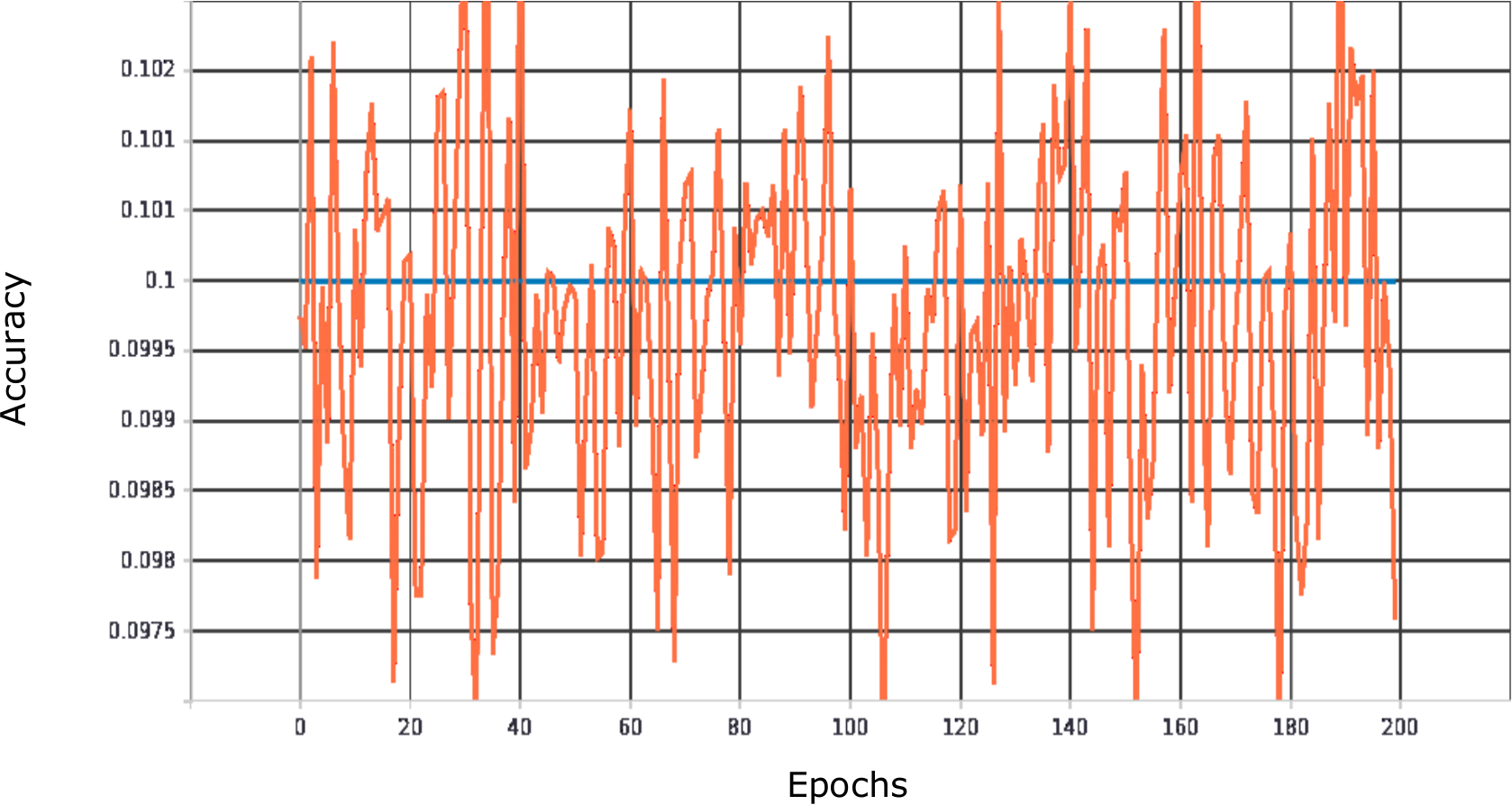}} 
	\subfloat[Accuracy of BO.]{\includegraphics[width=170pt, height=80pt]{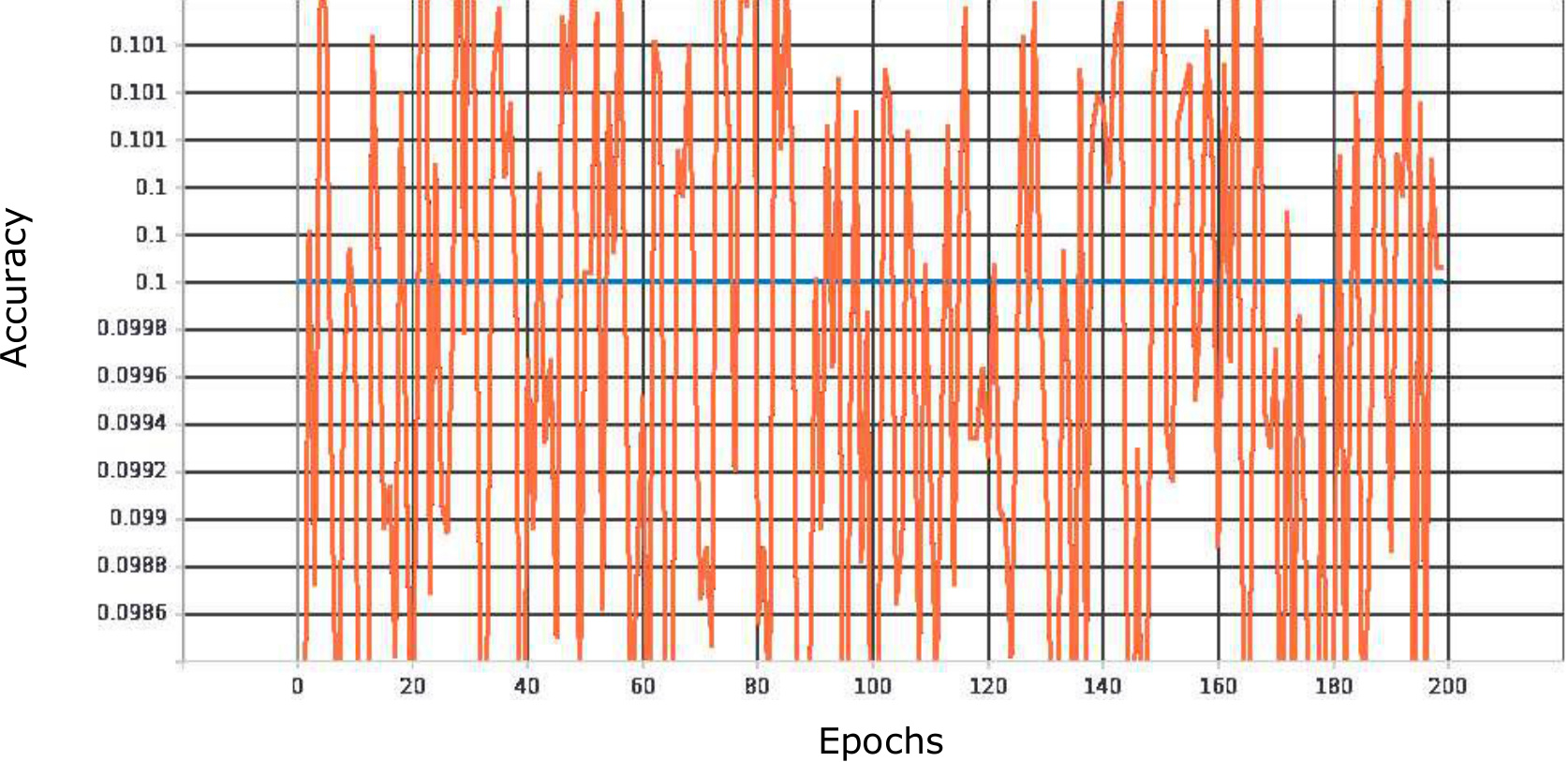}} \quad
	\subfloat[Loss of DNN-Tuner.]{\includegraphics[width=170pt, height=80pt]{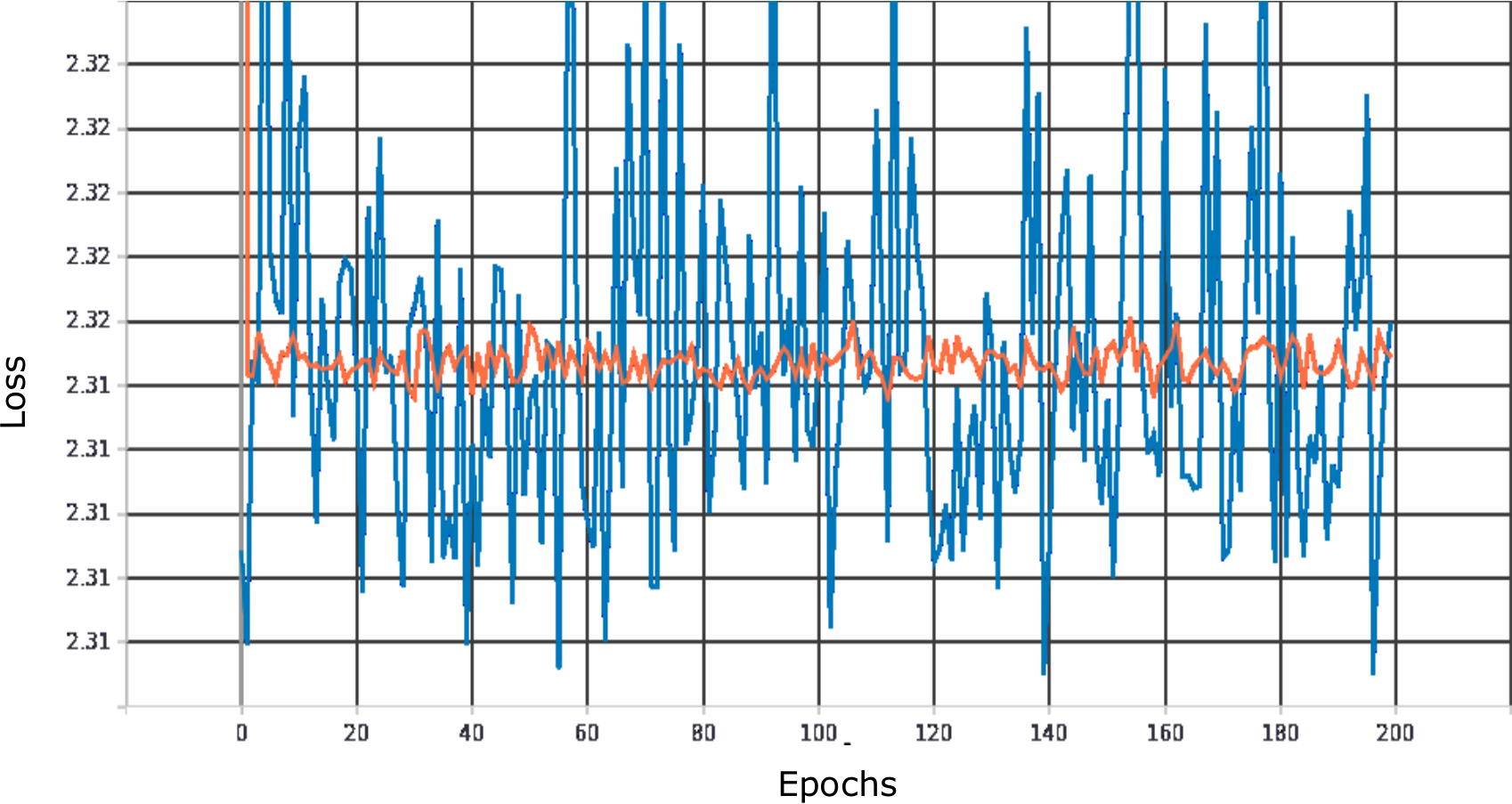}} 
	\subfloat[Loss of BO.]{\includegraphics[width=170pt, height=80pt]{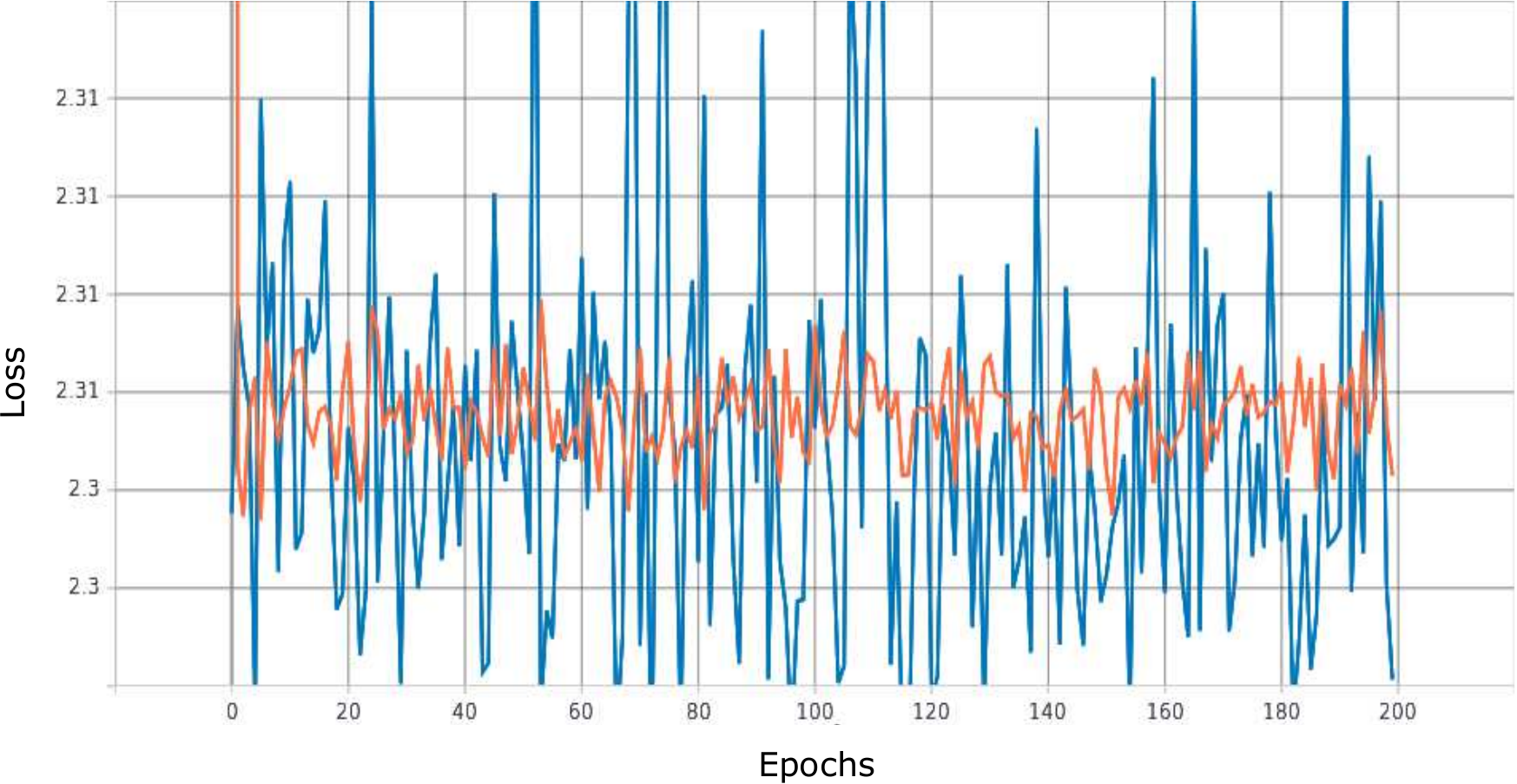}}
	\caption{Accuracy and loss in training (orange line) and validation (blue line) in the first step of both algorithms (DNN-Tuner and Bayesian Optimization).} \label{fig: acc-loss-1}
\end{figure}

\begin{figure}
	\centering
	\subfloat[Accuracy of DNN-Tuner.]{\includegraphics[width=170pt, height=80pt]{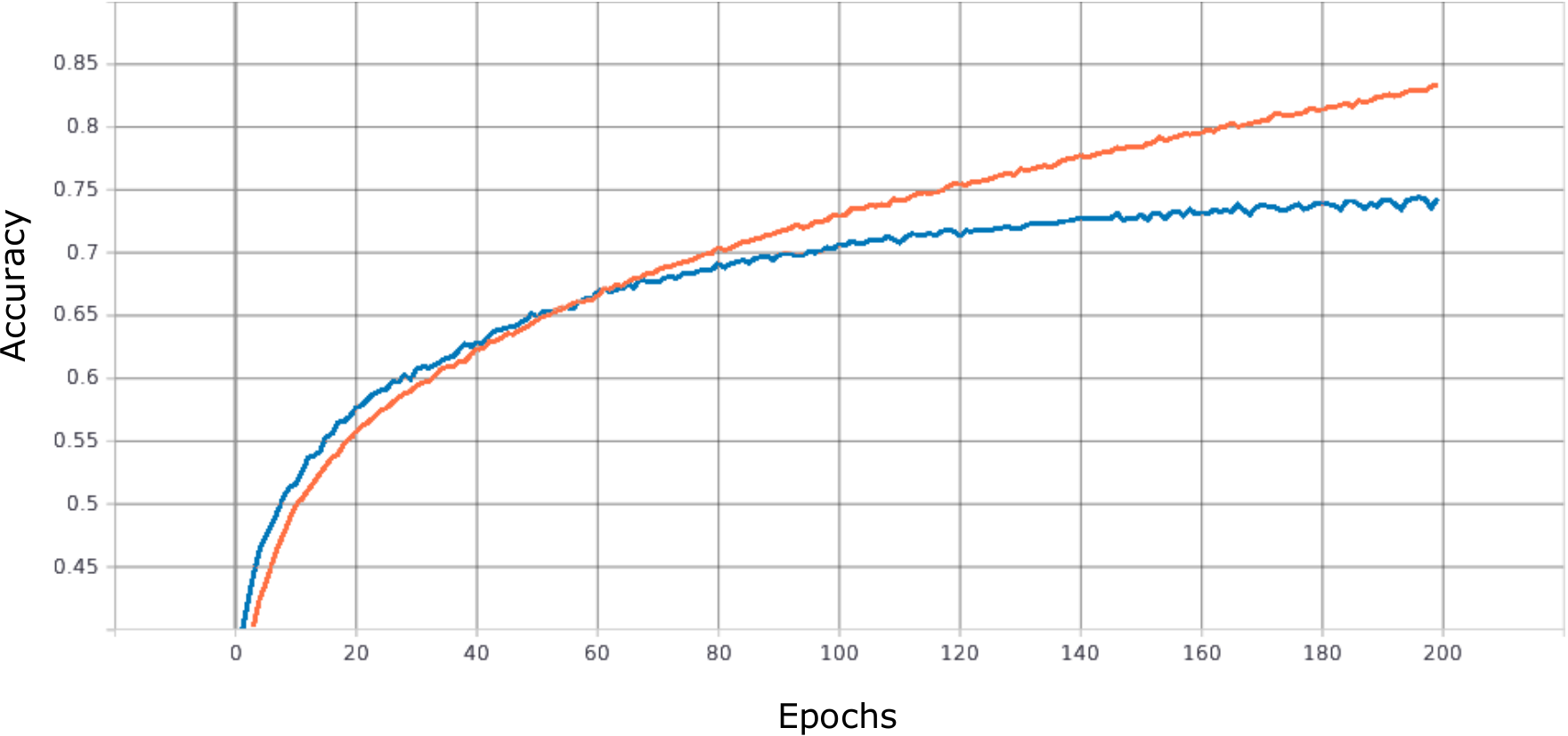}} 
	\subfloat[Accuracy of BO.]{\includegraphics[width=170pt, height=80pt]{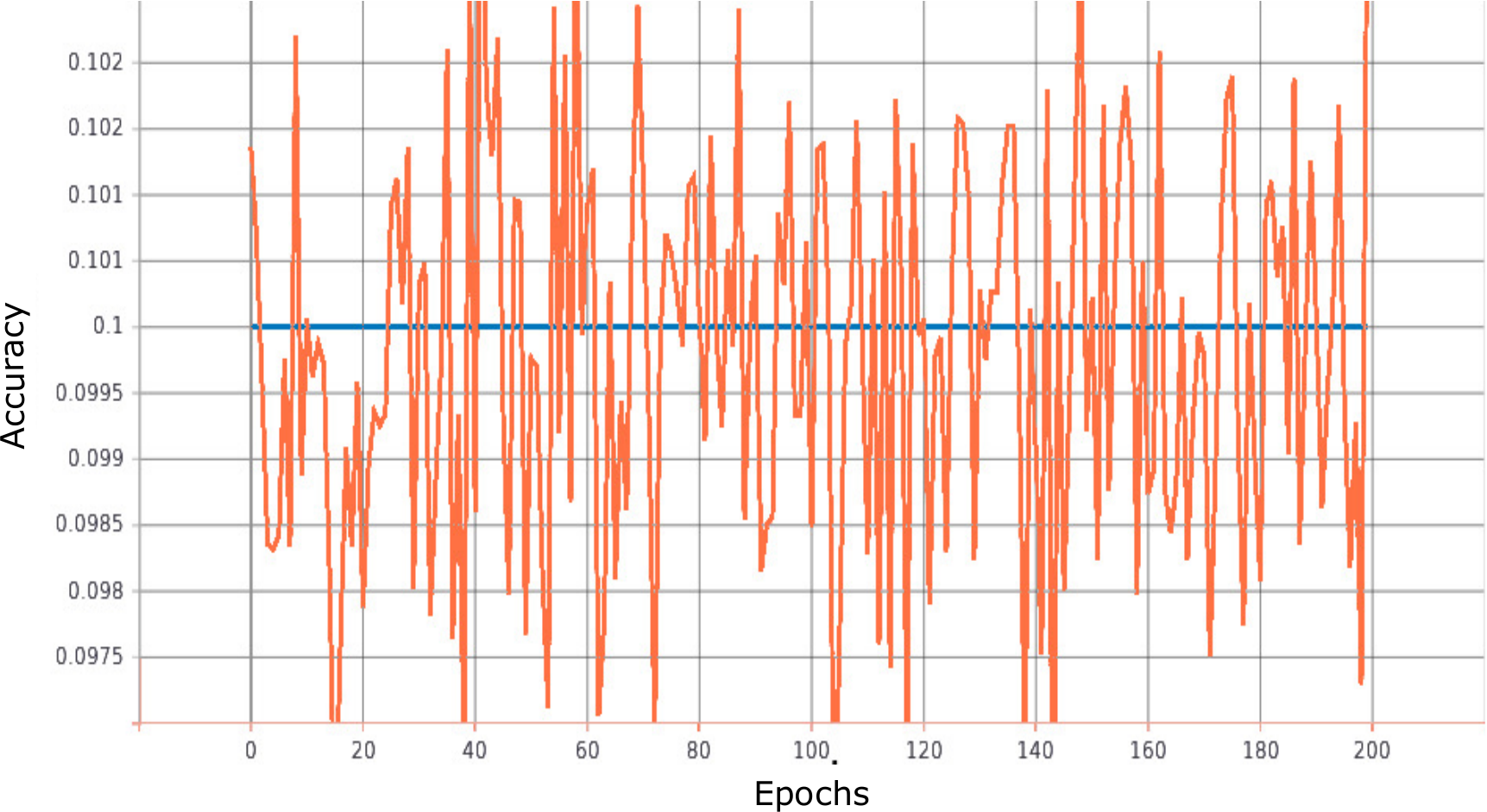}} \quad
	\subfloat[Loss of DNN-Tuner.]{\includegraphics[width=170pt, height=80pt]{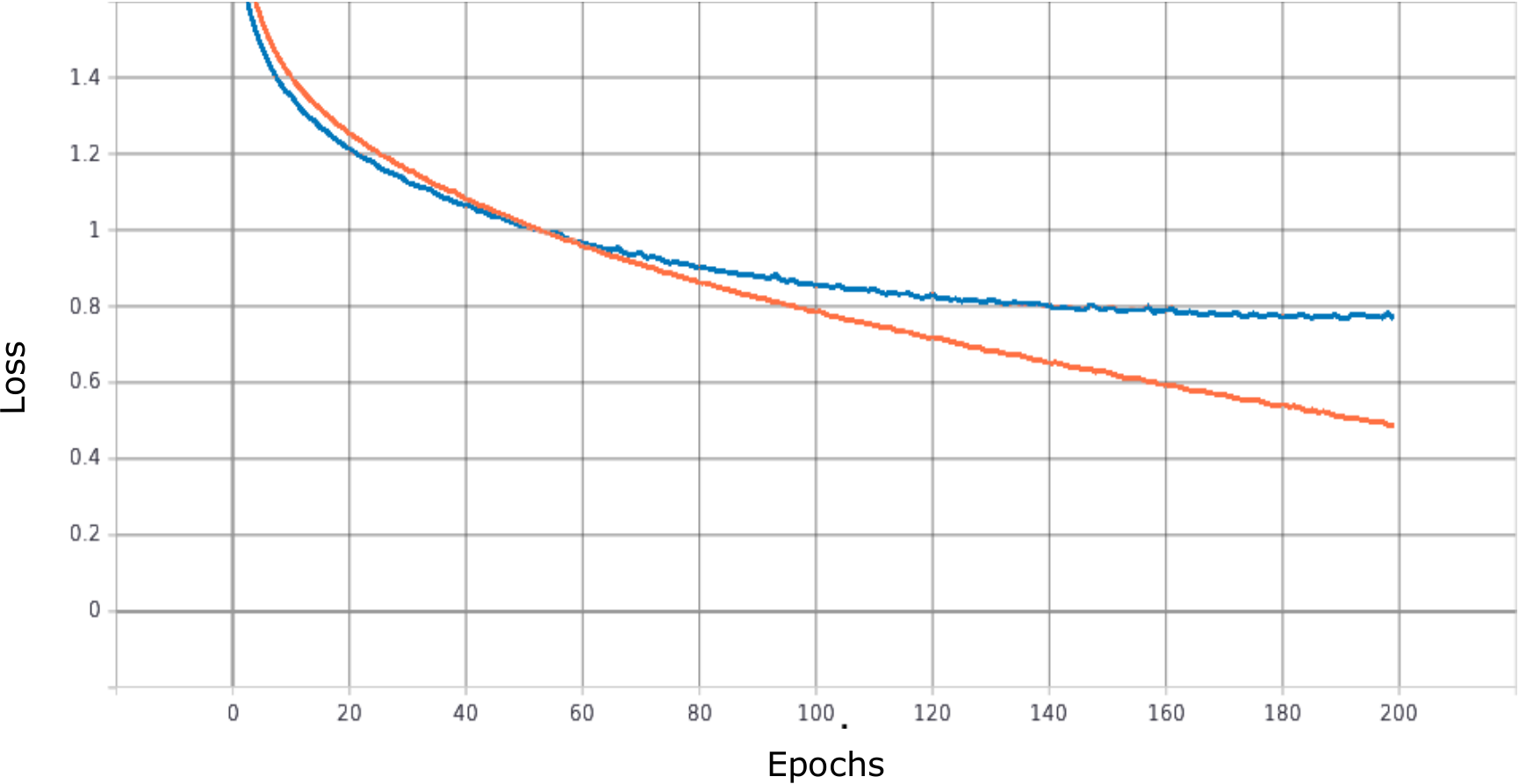}} 
	\subfloat[Loss of BO.]{\includegraphics[width=170pt, height=80pt]{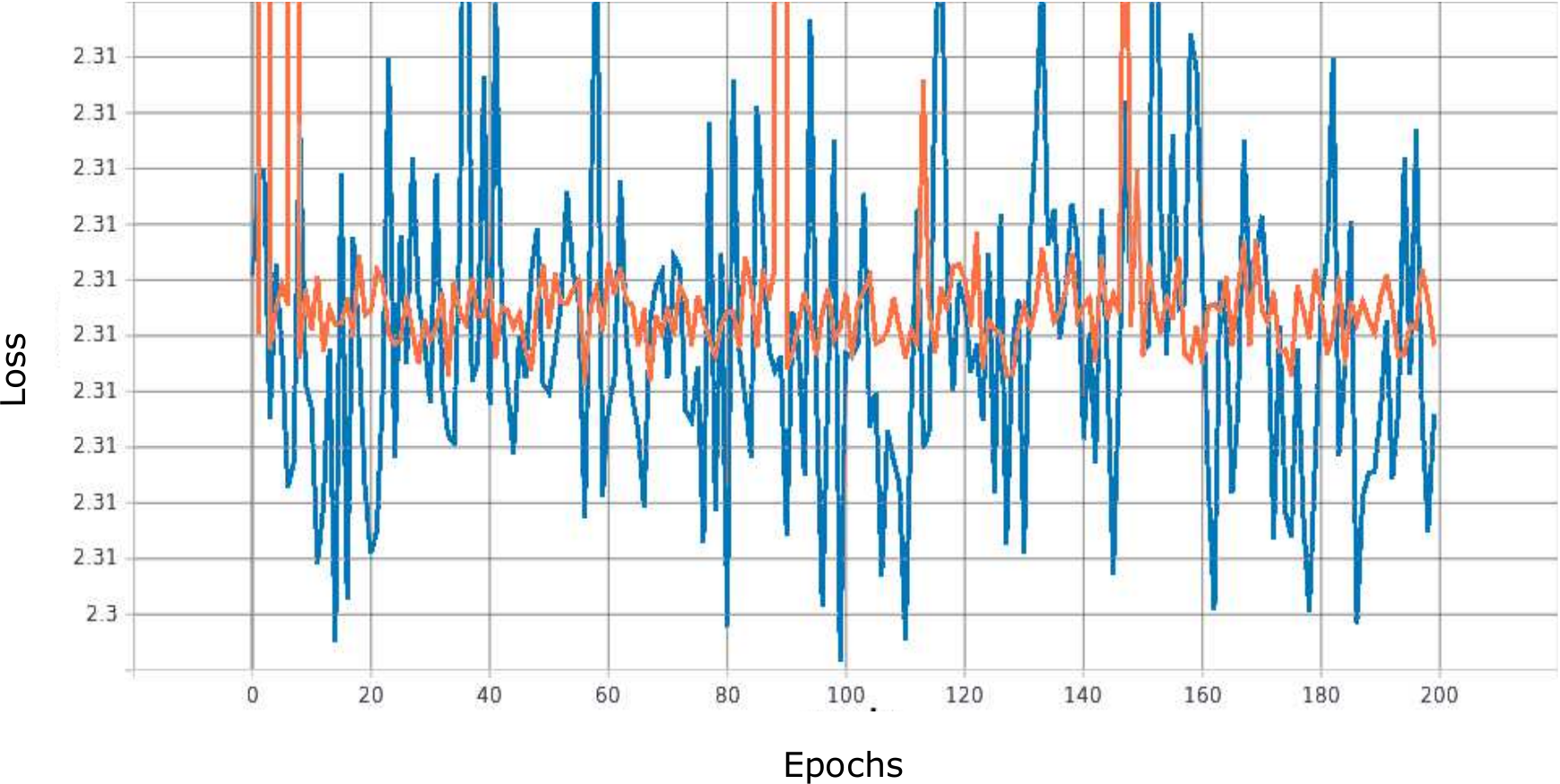}}
	\caption{Accuracy and loss in training (orange line) and validation (blue line) in the fourth step of both algorithms.} \label{fig: acc-loss-4}
\end{figure}

\begin{figure}
	\centering
	\subfloat[Accuracy of DNN-Tuner.]{\includegraphics[width=170pt, height=80pt]{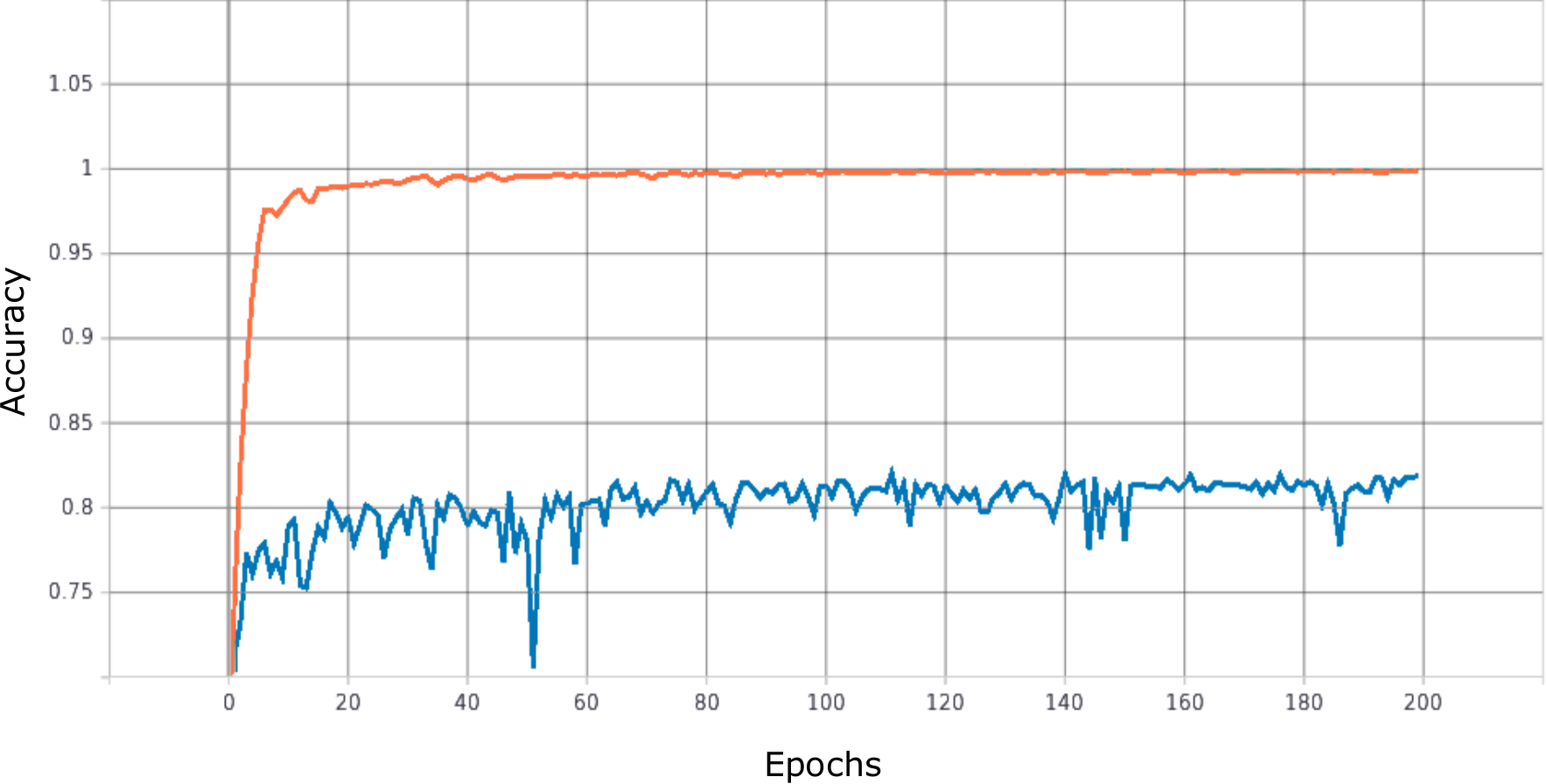}} 
	\subfloat[Accuracy of BO.]{\includegraphics[width=170pt, height=80pt]{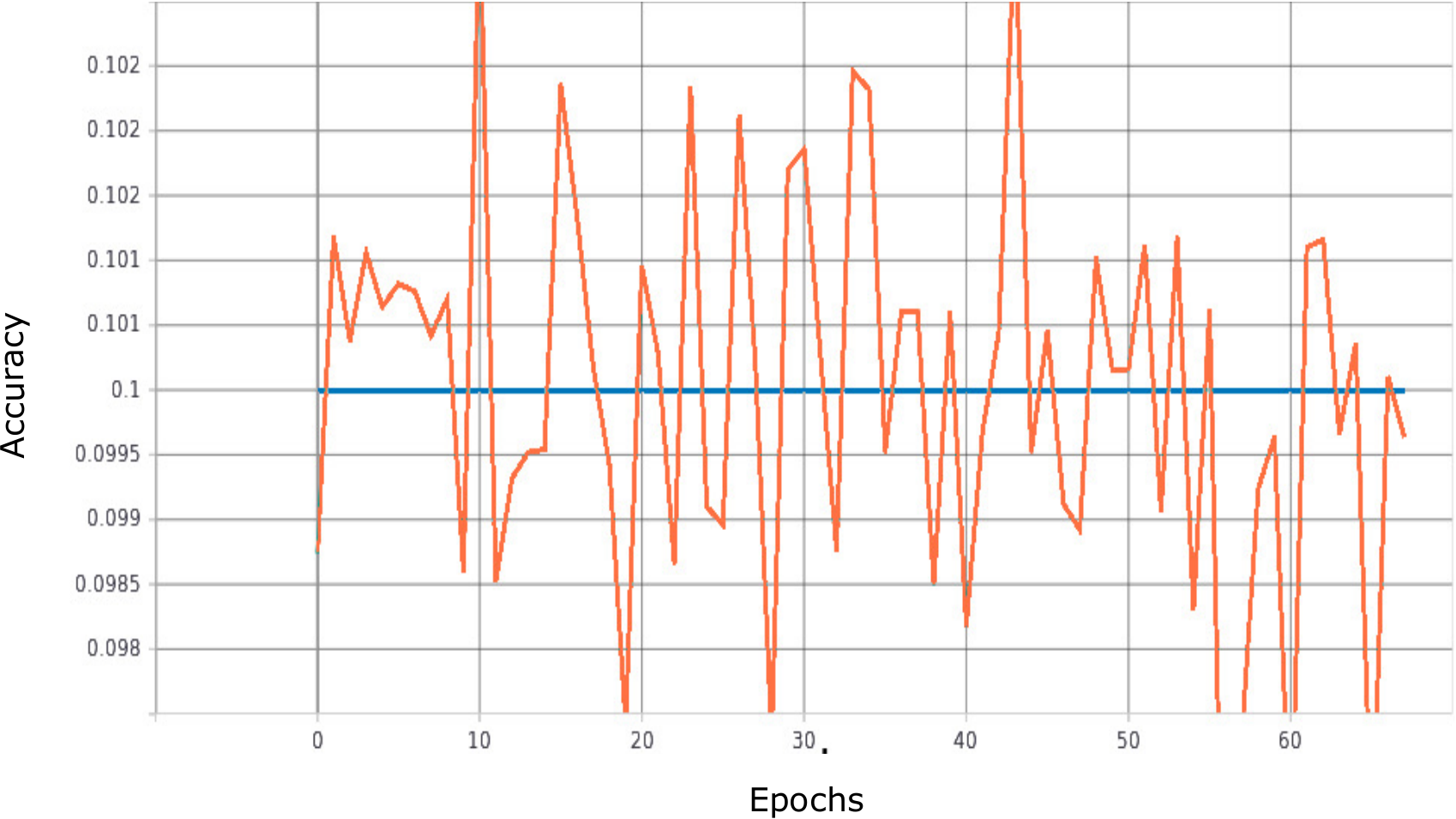}} \quad
	\subfloat[Loss of DNN-Tuner.]{\includegraphics[width=170pt, height=80pt]{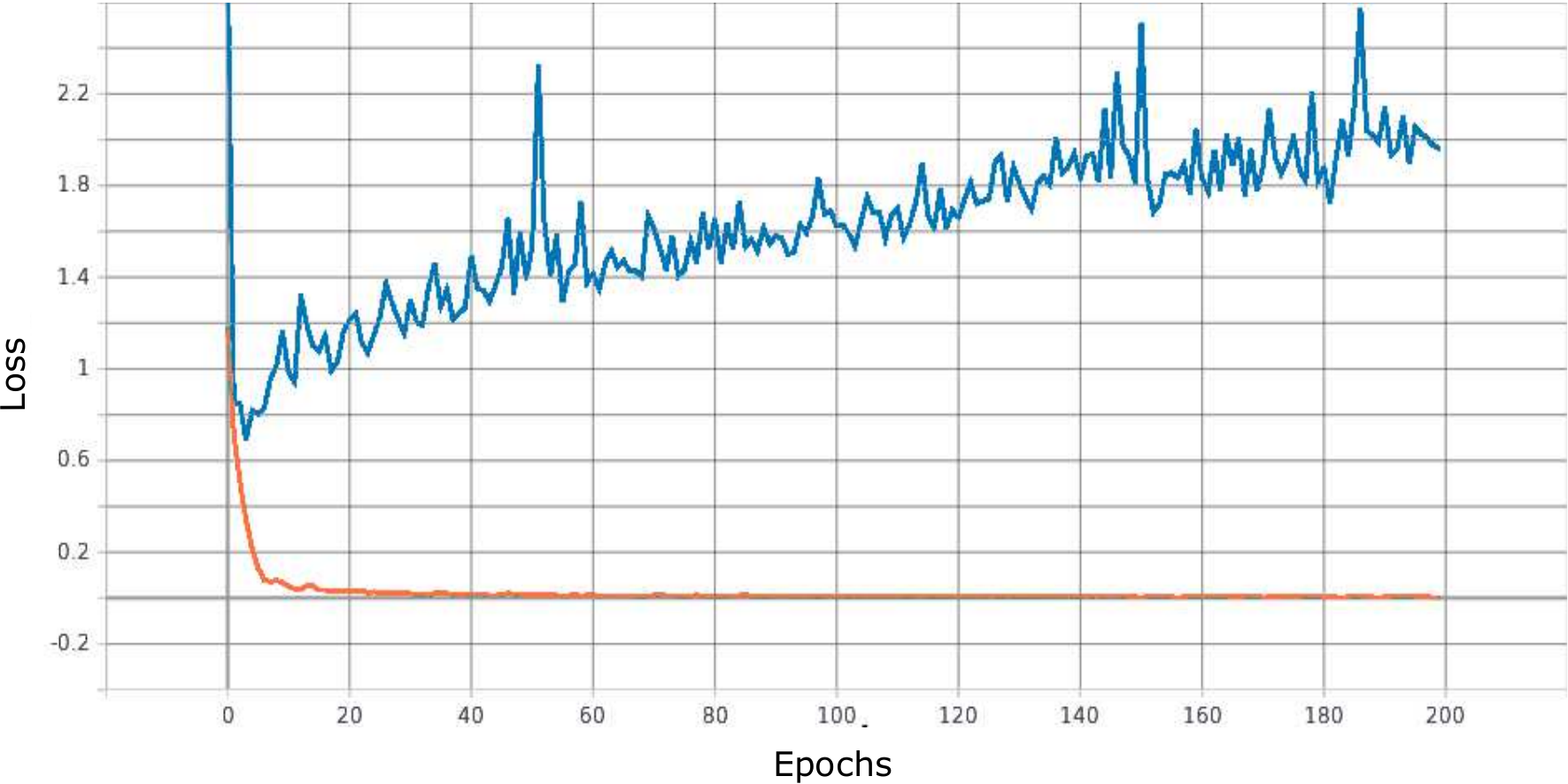}} 
	\subfloat[Loss of BO.]{\includegraphics[width=170pt, height=80pt]{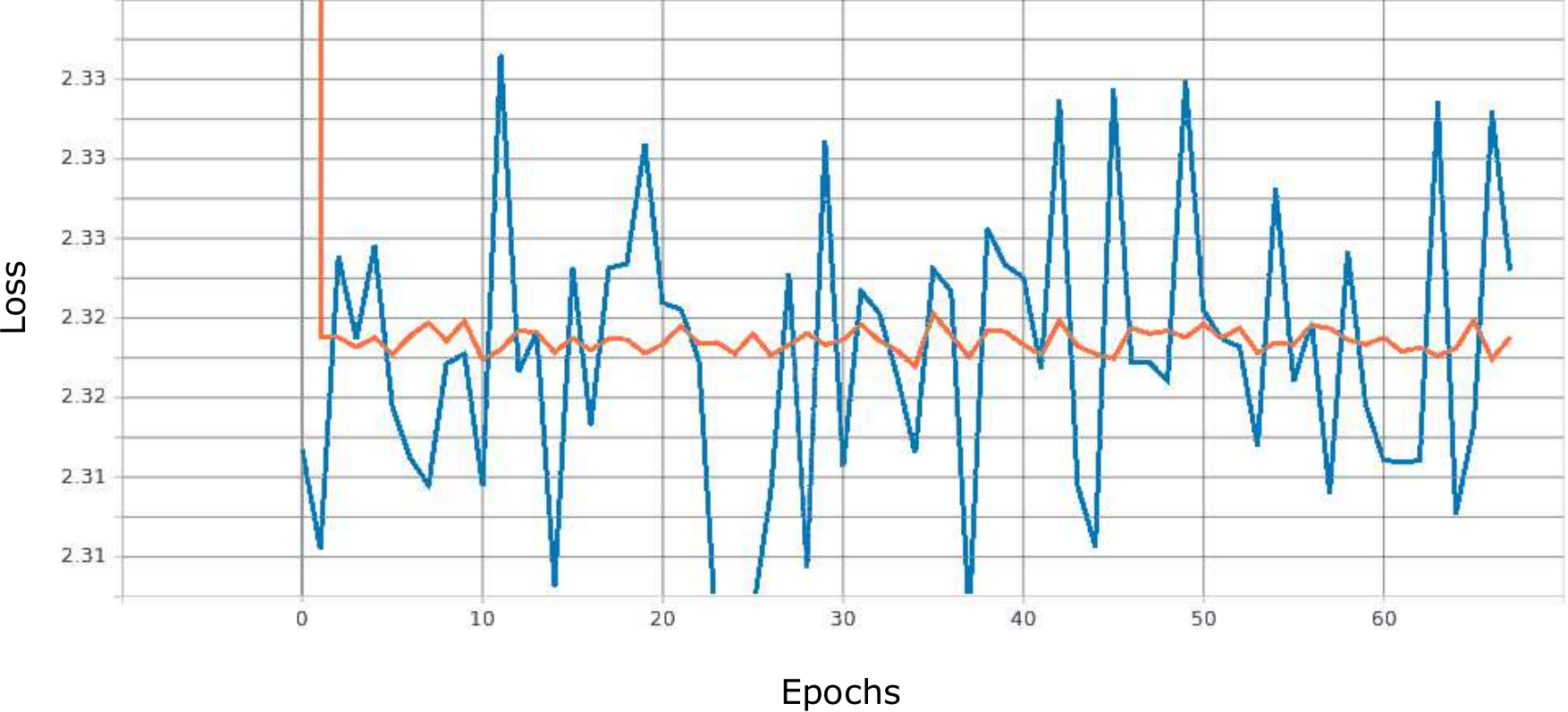}}
	\caption{Accuracy and loss in training (orange line) and validation (blue line) in the seventh step of both algorithms.} \label{fig: acc-loss-7}
\end{figure}

\begin{figure}
	\centering
	\subfloat[Accuracy of DNN-Tuner.]{\includegraphics[width=170pt, height=85pt]{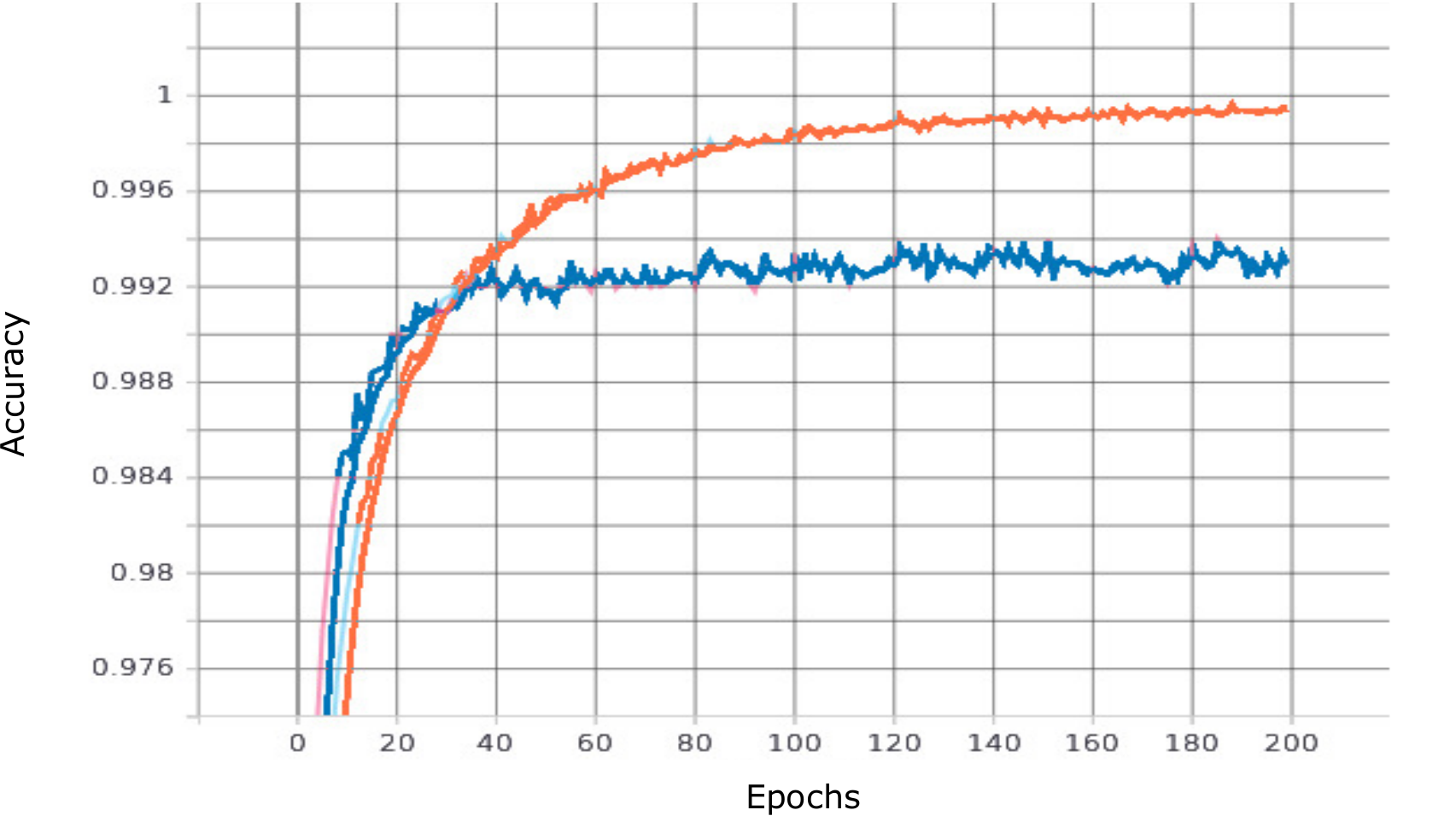}} 
	\subfloat[Accuracy of BO.]{\includegraphics[width=170pt, height=85pt]{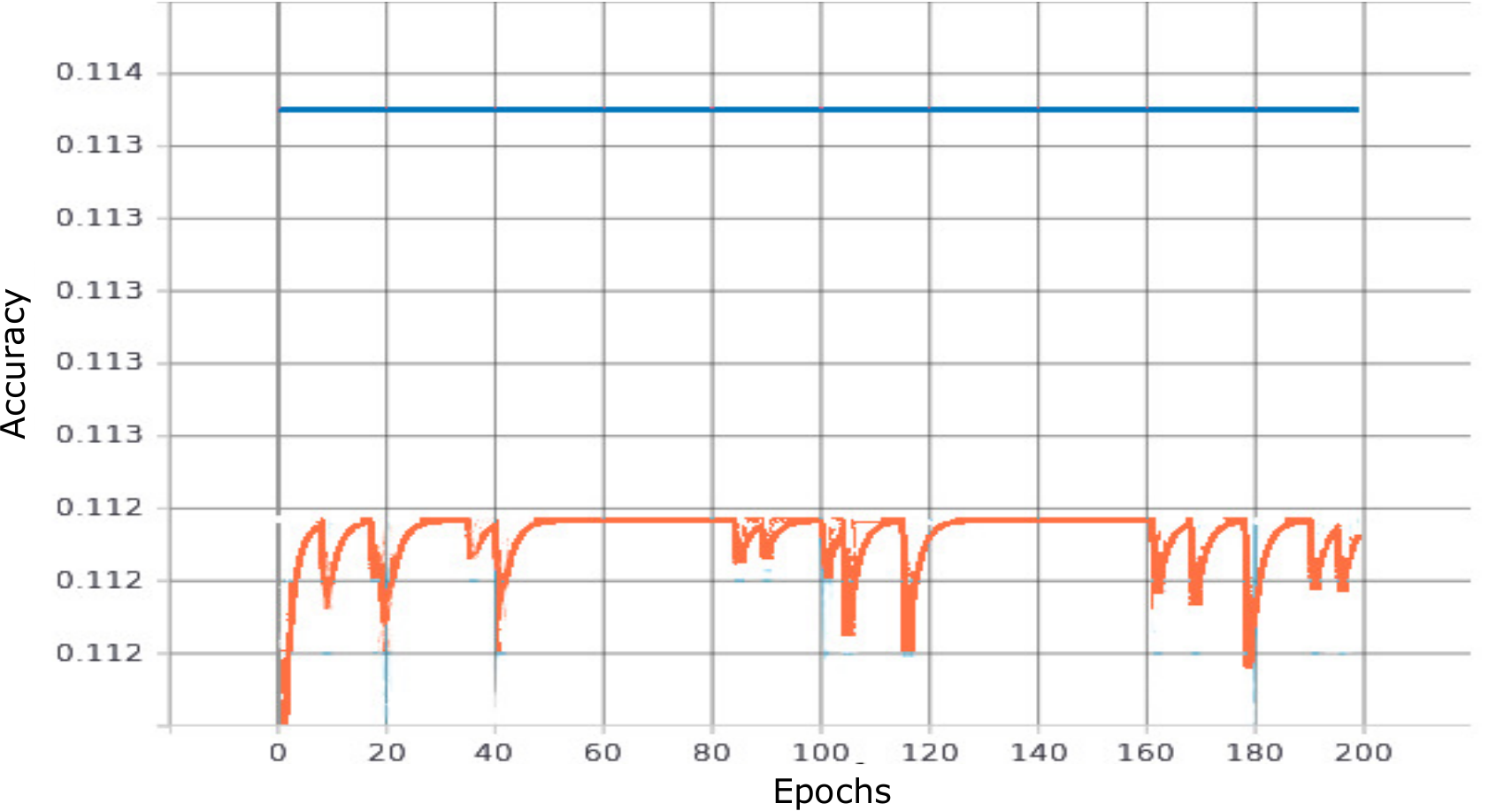}} 
	\caption{Accuracy and loss in training (orange line) and validation (blue line) of the seventh step of both algorithms.} \label{fig: acc-loss-mnist-7_1}
\end{figure}

\begin{figure}
	\centering
	\subfloat[Loss of DNN-Tuner.]{\includegraphics[width=170pt, height=85pt]{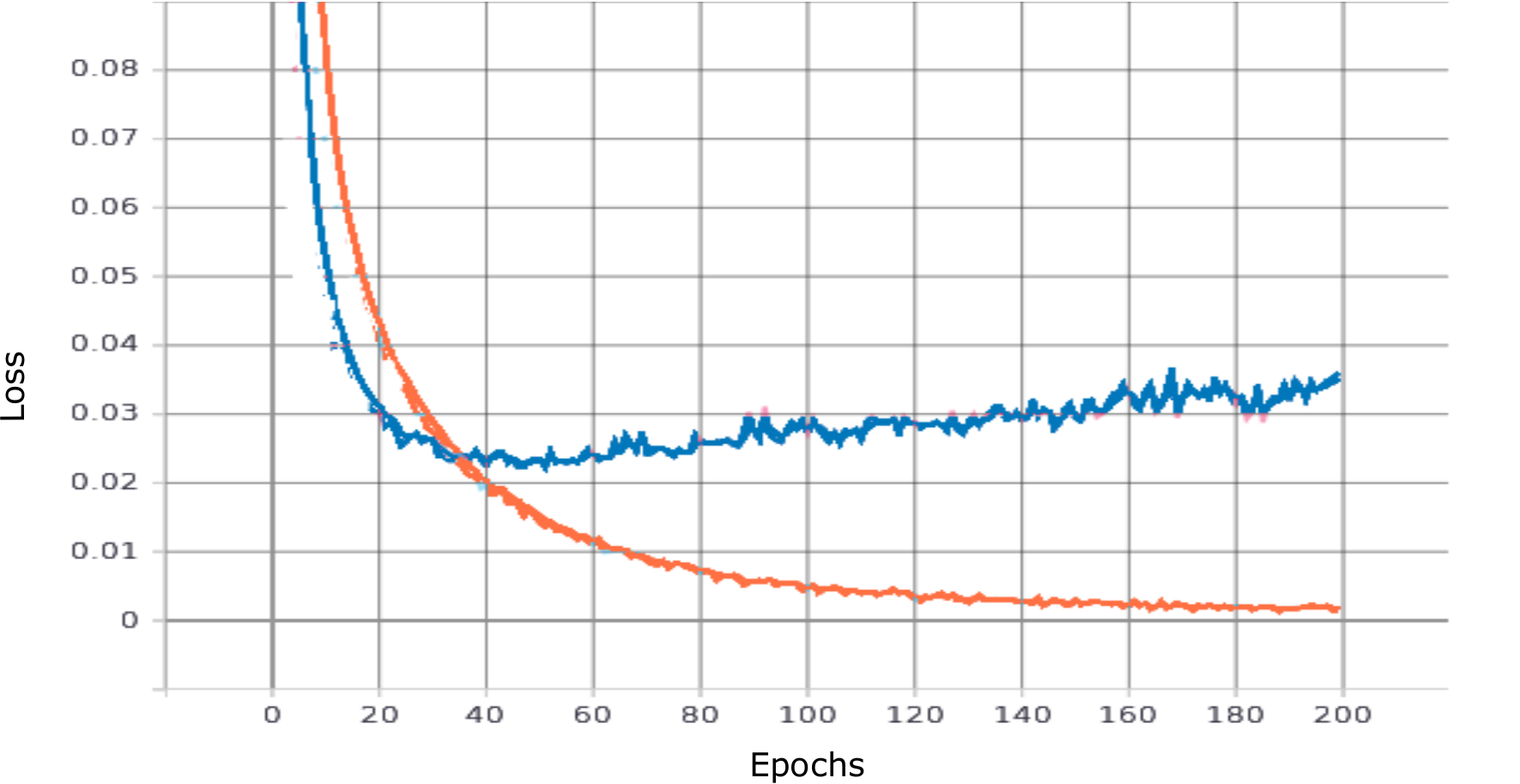}} 
	\subfloat[Loss of BO.]{\includegraphics[width=170pt, height=85pt]{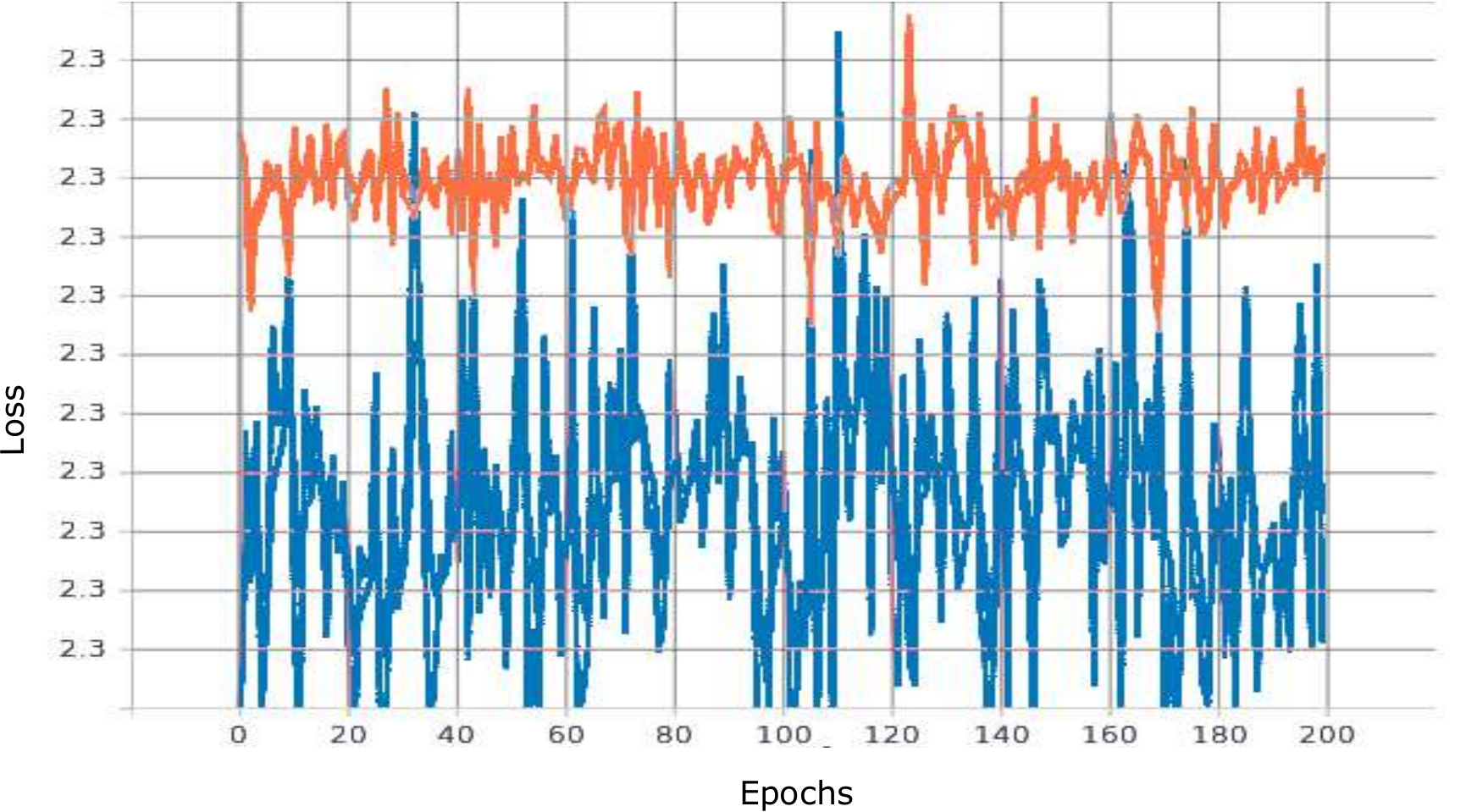}}
	\caption{Accuracy and loss in training (orange line) and validation (blue line) of the seventh step of both algorithms.} \label{fig: acc-loss-mnist-7_2}
\end{figure}

Figures \ref{fig: acc-loss-1} to \ref{fig: acc-loss-mnist-7_2} show accuracy and loss in training and validation phases of both algorithms (DNN-Tuner and BO). Both algorithms have completed seven optimization cycles. The neural network has been trained for 200 epochs and the dataset used are CIFAR-10 and MNIST.

Figures \ref{fig: acc-loss-1} show the difference in training and validation at the first cycle of optimization of DNN-Tuner and Bayesian Optimization respectively. You can observe that the trends are noisy because, in the first cycle, the hyper-parameters' values are chosen randomly.

Figures \ref{fig: acc-loss-4} show the difference in training and validation at the fourth cycle of optimization of DNN-Tuner and Bayesian Optimization respectively. You can observe that the trends of BO continue to be noisy, while the DNN-Tuner trends become less fluctuating with an accuracy of $\sim0.75$ and a loss of $\sim0.8$ on the validation set.

Figures \ref{fig: acc-loss-7} show the difference in training and validation at the seventh and last cycle of optimization of DNN-Tuner and Bayesian Optimization respectively. You can observe that the trends of BO still continues to be noisy, while the DNN-Tuner trends reach an accuracy of $\sim0.83$ and a loss of $\sim2$ on the validation set.

\FloatBarrier
Figures \ref{fig: acc-loss-mnist-7_1} and  \ref{fig: acc-loss-mnist-7_2} show the difference of training and validation at the seventh and last cycle of optimization of DNN-Tuner and Bayesian Optimization on the MNIST dataset, also with 200 training epochs. We can see that DNN-Tuner, reaches an accuracy $\sim0.993$ and a loss of $\sim0.04$ in the validation set unlike BO which reaches $\sim0.114$ and $\sim2.3$.

In terms of time, for the seven cycles of optimization with 200 training epochs on MNIST dataset: DNN-Tuner employed 4.75 hours while Bayesian Optimization 5.74 hours.

\FloatBarrier

\section{Conclusion and Future Work}
\label{conclusion}
DNNs have achieved extraordinary results but the number of hyper-parameters to be set has led to new challenges in the field of Artificial Intelligence.  
Given this problem and the ever-larger size of the networks, the standard approaches of hyper-parameters tuning like Grid and Random Search are no longer able to provide satisfactory results in an acceptable time. 
Inspired by the Bayesian Optimization (BO) technique, we have inserted expert knowledge into the tuning process, comparing our algorithm with BO in terms of quality of the results and time. 
The experiments show that it is possible to combine these approaches to obtain a better optimization of DNNs. The new algorithm implements tuning rules that work on the network structure and on the hyper-parameter' search space to drive the training of the network towards better results.  Some experiments were performed over standard datasets to demonstrate the improvement provided by DNN-Tuner. 
Future works will be focused on the implementation of the tuning rules with symbolic languages in order to achieve higher modularity, and interpretability (as well as explainability) for them.

\section*{Acknowledgements}
	The authors would like to acknowledge the project
	POR-FESR 2014-2020 "SUPER: Supercomputing Unified Platform - Emilia-Romagna". 
	This work was also supported by the National Group of Computing Science (GNCS-INDAM)".
	The first author is supported by a scholarship funded by Emilia-Romagna region, under POR FSE 2014-2020 program.

%
%
%
\bibliographystyle{splncs04}
\bibliography{bibl}
\end{document}